\PassOptionsToPackage{unicode}{hyperref}
\PassOptionsToPackage{hyphens}{url}
\PassOptionsToPackage{dvipsnames,svgnames,x11names}{xcolor}
\documentclass[
]{article}

\usepackage{amsmath,amssymb}
\usepackage{iftex}
\ifPDFTeX
  \usepackage[T1]{fontenc}
  \usepackage[utf8]{inputenc}
  \usepackage{textcomp} 
\else 
  \usepackage{unicode-math}
  \defaultfontfeatures{Scale=MatchLowercase}
  \defaultfontfeatures[\rmfamily]{Ligatures=TeX,Scale=1}
\fi
\usepackage{lmodern}
\ifPDFTeX\else  
\fi
\IfFileExists{upquote.sty}{\usepackage{upquote}}{}
\IfFileExists{microtype.sty}{
  \usepackage[]{microtype}
  \UseMicrotypeSet[protrusion]{basicmath} 
}{}
\makeatletter
\@ifundefined{KOMAClassName}{
  \IfFileExists{parskip.sty}{%
    \usepackage{parskip}
  }{
    \setlength{\parindent}{0pt}
    \setlength{\parskip}{6pt plus 2pt minus 1pt}}
}{
  \KOMAoptions{parskip=half}}
\makeatother
\usepackage{xcolor}
\ifx\paragraph\undefined\else
  \let\oldparagraph\paragraph
  \renewcommand{\paragraph}[1]{\oldparagraph{#1}\mbox{}}
\fi
\ifx\subparagraph\undefined\else
  \let\oldsubparagraph\subparagraph
  \renewcommand{\subparagraph}[1]{\oldsubparagraph{#1}\mbox{}}
\fi

\usepackage{longtable,booktabs,array}
\usepackage{calc} 
\usepackage{etoolbox}
\makeatletter
\patchcmd\longtable{\par}{\if@noskipsec\mbox{}\fi\par}{}{}
\makeatother
\IfFileExists{footnotehyper.sty}{\usepackage{footnotehyper}}{\usepackage{footnote}}
\makesavenoteenv{longtable}
\usepackage{graphicx}
\makeatletter
\def\maxwidth{\ifdim\Gin@nat@width>\linewidth\linewidth\else\Gin@nat@width\fi}
\def\maxheight{\ifdim\Gin@nat@height>\textheight\textheight\else\Gin@nat@height\fi}
\makeatother
\setkeys{Gin}{width=\maxwidth,height=\maxheight,keepaspectratio}
\makeatletter
\def\fps@figure{htbp}
\makeatother
\newlength{\cslhangindent}
\newlength{\csllabelwidth}
\newlength{\cslentryspacingunit} 
\newenvironment{CSLReferences}[2] 
 {
  \setlength{\parindent}{0pt}
  \ifodd #1
  \let\oldpar\par
  \def\par{\hangindent=\cslhangindent\oldpar}
  \fi
  \setlength{\parskip}{#2\cslentryspacingunit}
 }%
 {}
\usepackage{calc}

\usepackage{arxiv}
\usepackage{orcidlink}
\usepackage{amsmath}
\usepackage[T1]{fontenc}
\makeatletter
\@ifpackageloaded{caption}{}{\usepackage{caption}}
\AtBeginDocument{%
\ifdefined\contentsname
  \renewcommand*\contentsname{Table of contents}
\else
  \newcommand\contentsname{Table of contents}
\fi
\ifdefined\listfigurename
  \renewcommand*\listfigurename{List of Figures}
\else
  \newcommand\listfigurename{List of Figures}
\fi
\ifdefined\listtablename
  \renewcommand*\listtablename{List of Tables}
\else
  \newcommand\listtablename{List of Tables}
\fi
\ifdefined\figurename
  \renewcommand*\figurename{Figure}
\else
  \newcommand\figurename{Figure}
\fi
\ifdefined\tablename
  \renewcommand*\tablename{Table}
\else
  \newcommand\tablename{Table}
\fi
}
\@ifpackageloaded{float}{}{\usepackage{float}}
\floatstyle{ruled}
\@ifundefined{c@chapter}{\newfloat{codelisting}{h}{lop}}{\newfloat{codelisting}{h}{lop}[chapter]}
\floatname{codelisting}{Listing}

\makeatother
\makeatletter
\@ifpackageloaded{caption}{}{\usepackage{caption}}
\@ifpackageloaded{subcaption}{}{\usepackage{subcaption}}
\makeatother
\makeatletter
\@ifpackageloaded{tcolorbox}{}{\usepackage[skins,breakable]{tcolorbox}}
\makeatother
\makeatletter
\@ifundefined{shadecolor}{\definecolor{shadecolor}{rgb}{.97, .97, .97}}
\makeatother
\ifLuaTeX
  \usepackage{selnolig}  
\fi
\IfFileExists{bookmark.sty}{\usepackage{bookmark}}{\usepackage{hyperref}}
\IfFileExists{xurl.sty}{\usepackage{xurl}}{} 
\hypersetup{
  pdftitle={SurveyLM: a platform to explore emerging value perspectives in Augmented Language Models' behaviors},
  pdfauthor={Steve J. Bickley, Ho Fai Chan, Bang Dao, Benno Torgler, Son Tran},
  pdfkeywords={Social Science, Social AI, Large Language
Models, Augmented Language Models, Alignment Research, Survey
Method, Experimental Method},
  colorlinks=true,
  linkcolor={blue},
  filecolor={Maroon},
  citecolor={Blue},
  urlcolor={Blue},
  pdfcreator={LaTeX via pandoc}}

\renewcommand{\today}{01.08.2023}
\newcommand{\runninghead}{A Preprint }
\renewcommand{\runninghead}{Panalogy Lab Technical Report 2023-001 }
\title{SurveyLM: a platform to explore emerging value perspectives in
Augmented Language Models' behaviors}
\author{
\textbf{Steve J. Bickley, Ho Fai Chan, Bang Dao, Benno Torgler, Son
Tran}\\\\Panalogy
Lab\\Brisbane,\ 4000\\\href{mailto:steven.bickley@panalogy-lab.com}{steven.bickley@panalogy-lab.com}}
\date{01.08.2023}
\begin{document}
\maketitle
\begin{abstract}
This white paper presents our work on SurveyLM, a platform for analyzing
augmented language models' (ALMs) emergent alignment behaviors through
their dynamically evolving attitude and value perspectives in complex
social contexts. Social Artificial Intelligence (AI) systems, like ALMs,
often function within nuanced social scenarios where there is no
singular correct response, or where an answer is heavily dependent on
contextual factors, thus necessitating an in-depth understanding of
their alignment dynamics. To address this, we apply survey and
experimental methodologies, traditionally used in studying social
behaviors, to evaluate ALMs systematically, thus providing unprecedented
insights into their alignment and emergent behaviors. Moreover, the
SurveyLM platform leverages the ALMs' own feedback to enhance survey and
experiment designs, exploiting an underutilized aspect of ALMs, which
accelerates the development and testing of high-quality survey
frameworks while conserving resources. Through SurveyLM, we aim to shed
light on factors influencing ALMs' emergent behaviors, facilitate their
alignment with human intentions and expectations, and thereby contribute
to the responsible development and deployment of advanced social AI
systems. This white paper underscores the platform's potential to
deliver robust results, highlighting its significance to alignment
research and its implications for future social AI systems.
\end{abstract}
{\bfseries \emph Keywords}
\def\sep{\textbullet\ }
Social Science \sep Social AI \sep Large Language Models \sep Augmented
Language Models \sep Alignment Research \sep Survey Method \sep 
Experimental Method

\ifdefined\Shaded\renewenvironment{Shaded}{\begin{tcolorbox}[enhanced, borderline west={3pt}{0pt}{shadecolor}, interior hidden, breakable, boxrule=0pt, frame hidden, sharp corners]}{\end{tcolorbox}}\fi

\newpage{}

\hypertarget{sec-motivation}{%
\section{Motivation}\label{sec-motivation}}

We built SurveyLM with the following motivations in mind:

\begin{itemize}
\item
  Artificial Intelligence (AI) alignment in complex social context is
  \emph{important}, especially when AI systems make decisions or assist
  people in making decisions in complex social settings where there may
  be no true ``right'' answer or where an answer is heavily dependent on
  contextual factors and future uncertainties (specific information in
  real-world situations).
\item
  The application of survey and experimental methodologies, which are
  extensively utilized to study social behaviors, can greatly enhance
  the effectiveness of language model evaluation frameworks. These
  approaches can provide valuable insights into the behavior and
  alignment of the models, particularly when implemented in a
  \emph{systematic} manner.
\item
  Augmented language models (ALMs) are AI systems that currently have
  the most advanced general reasoning capability and are increasingly
  functioning within diverse and intricate social scenarios, exhibiting
  emergent behaviors that were previously unexpected. Therefore, it is
  desirable to apply survey and experimental methodologies to not only
  unravel the factors influencing these emergent behaviors, but also to
  ensure these ALMs are adapting and aligning with human intentions and
  expectations, thus contributing to the responsible development and
  deployment of advanced AI systems.
\item
  Being able to learn from ALM's feedback to \emph{improve}
  survey/experiment designs is a great and underexplored application of
  ALMs, thus helping researchers to construct high quality survey
  frameworks at a fraction of the resources and time that would be
  otherwise required (past approach where humans have been doing
  everything).
\item
  SurveyLM aims to be the first platform to deliver robust results for
  the above value propositions.
\end{itemize}

\hypertarget{sec-intro}{%
\section{Introduction}\label{sec-intro}}

Our new platform, SurveyLM, aims to explore emerging value perspectives
in ALMs (Mialon et al., 2023) through decision premises and
contextuality in complex social settings. We chose to focus on complex
social context because these environments present a wide array of
situations and (social) dilemmas that help assessing the multifaceted
and nuanced value judgments that ALMs have to make in the real world. As
a result, we can better understand their decision-making mechanisms and
the underlying value systems they rely on.

Since our focus is on complex social decision situations that have no
obvious binary classification of being totally right or wrong, we adopt
an approach that enables social science researchers to systematically
evaluate ALMs' decision dynamics in a diverse range of social contexts
using different measures, instead of benchmarking ALMs capabilities
against a simplified set of sequential decision making tasks often used
in the field of AI (Shridhar et al., 2021). While these tasks provide a
clear context to evaluate whether an ALM is right or wrong in making
some decisions, or in carrying out some specific tasks, they are far
from being adequate to capture real-world decision processes that social
science researchers are concerned with.

Our approach to exploring ALM's behaviors, and their value perspectives,
is based on presenting the models with complex sets of survey questions
and experimental settings, and then analyzing their responses in various
social contexts. We opted for a survey-based approach as it has been
proven to be an effective method for probing and value elicitation of
ALMs (Arora et al., 2023; Binz \& Schulz, 2023). Human value elicitation
has always been an important research avenue (Haerpfer et al., 2021;
Hofstede, 2005), as an individual's identity and values often manifest
in the choices and decisions they make. Given the increasing role of
ALMs as agents interacting with humans who act as principals (i.e.,
principal-agent relationship in nature), it becomes crucial to
comprehend the underlying values of these models and their implications
for recommendations and decision-making processes, which may or may not
align with those of human principals. By allowing the models to respond
to a series of contextualized questions, we can gain valuable insights
into how they navigate ethical scenarios and prioritize differing values
for different principal profiles and characteristics (e.g., age, gender,
ethnicity, country of residence, etc.).

The first key aspect we address is the question of why understanding
ALM's value perspectives matters in alignment research. By examining the
context-dependent values that drive ALMs' behaviors, we gain valuable
insights into the ethical and moral frameworks and traces that shape
their decision-making processes (as well as our own). These insights can
help us align ALMs with (context-specific) human values and promote
responsible AI development (Tamkin et al., 2021).

Defining measures of values is another crucial aspect of our platform,
SurveyLM. We recognize that values are multifaceted and subjective,
making their quantification challenging. Human values, as well as AI
alignment to these values, evolve over time and iterations,
necessitating a continuously adaptive approach. Through rigorous
research and analysis, we aim to develop and translate robust
methodologies to define and measure values in the context of ALMs, and
to discern trends and shifts in ALMs value systems and alignment. This
will provide a solid foundation for studying and understanding their
behavior patterns in complex social settings.

To effectively explore emerging value perspectives in ALMs' behaviors,
we employ state-of-the-art ALM simulation techniques. By simulating ALMs
in realistic scenarios, we can observe their decision-making processes
and identify any evolving biases or value systems. This knowledge is
crucial for staying ahead of potential risks and ensuring the
responsible development of AI technologies.

The potential applications of our platform extend beyond the fields of
behavioral economics, cognitive psychology, management or market
research. We envision leveraging the insights gained to help drive
social AI alignment and design in research, industry and community
alike. By aligning AI systems with human values, we can (co-)create
social AI systems that complement and enhance lives, benefiting society
as a whole.

Overall, our framework proposed provides an effective and user-friendly
platform for researchers to interact with GPT models. It upholds crucial
principles such as user-first orientation, systematic and consistent
operations, privacy respect, and safety in worst-case scenarios.
Furthermore, it features robust components including a simple front-end
interface, a real-time metrics monitor, secure databases, efficient
agents, and a smart requests manager, all engineered to make complex
survey research a breeze. We hope that this platform holds the potential
to revolutionize how researchers engage with GPT models, offering a
unique blend of flexibility, efficiency, and reliability.

In conclusion, our platform SurveyLM offers a unique opportunity to
explore and understand the emerging value perspectives from ALMs'
behaviors. By addressing the questions of why values matter, defining
measures of values, and simulating ALMs in complex social settings, we
can generate insights that will help shape the future of AI development.
With a focus on social AI alignment and design, we aim to create
long-lived, complex social AI agents and systems that align with human
values, promoting a responsible and beneficial integration of AI into
industry and society.

\hypertarget{sec-paltform-design}{%
\section{Platform Design}\label{sec-paltform-design}}

The design of our platform is guided by several key principles centered
around efficiency, flexibility, and user privacy.

\textbf{Put Researchers First:} This principle is the cornerstone of our
design framework. The goal is to make the researcher's interaction with
the platform as intuitive and seamless as possible, reducing the
complexity of the underlying operations. From parameter configuration to
document and data processing, we aim to create an environment that
empowers the user to focus solely on their research questions and
outcomes. To achieve this, complex prompt engineering and API request
handling will be handled internally by the platform, abstracting the
process and reducing the technical knowledge necessary to operate the
tool effectively.

\textbf{Respect Research Logic:} We design the platform in such a way
that, despite layers of operational complexity, keeps the research logic
intact by translating parameter configurations into prompt structures
that ensure LLMs' responses would be shaped by only those
configurations. This is achieved by giving researchers freedom in
defining transparently what comes into and what comes out of a
simulation process. Whereas, the platform focuses on building optimal
prompt structure and request handling by using researchers' inputs with
minimal additional content needed to interface consistently with LLMs.

\textbf{Be Systematic Yet Flexible:} A systematic approach to parameter
configuration is crucial to accommodate a diverse range of research
scenarios. With the flexibility to modify parameters as per the needs of
the study, the platform allows researchers to experiment and customize
the models' responses, thereby facilitating a broader spectrum of
research.

\textbf{Embrace Consistency and Dynamism:} Consistency is paramount for
any research work. As such, the platform will ensure that the GPT models
produce consistent responses to survey questions in terms of answer
formats and compliance with other instructions. At the same time, we
recognize the evolving nature of AI, incorporating the dynamic aspect of
learning and growth in AI models to our consistency ethos. Balancing
these elements enhances not just the reproducibility, but also the
adaptive validity and reliability of our research findings.

\textbf{Respect Privacy and Data Control:} With the rise of data
breaches, privacy protection is a critical concern for users. On our
platform, we uphold the principle of data control, retaining user data
exclusively until such time as the user decides to delete it. Upon
deletion, all data will be purged except for the basic user credentials.
This ensures that users' privacy and data security are upheld without
compromising the platform's functionality.

\textbf{Safe Worse Case:} The platform will provide robustness against
system errors, particularly those caused by API rate limits. Should an
error occur, the platform will ensure that all simulation data leading
up to the error are preserved. Additionally, any uncompleted operations
will also be recorded and sent to the user for further decision making.
This principle allows for seamless recovery, ensuring users' work is
never lost and fostering trust in the platform's reliability.

Based on these principles, SurveyLM was built with several components as
shown in Figure 1.

\begin{figure}

{\centering \includegraphics{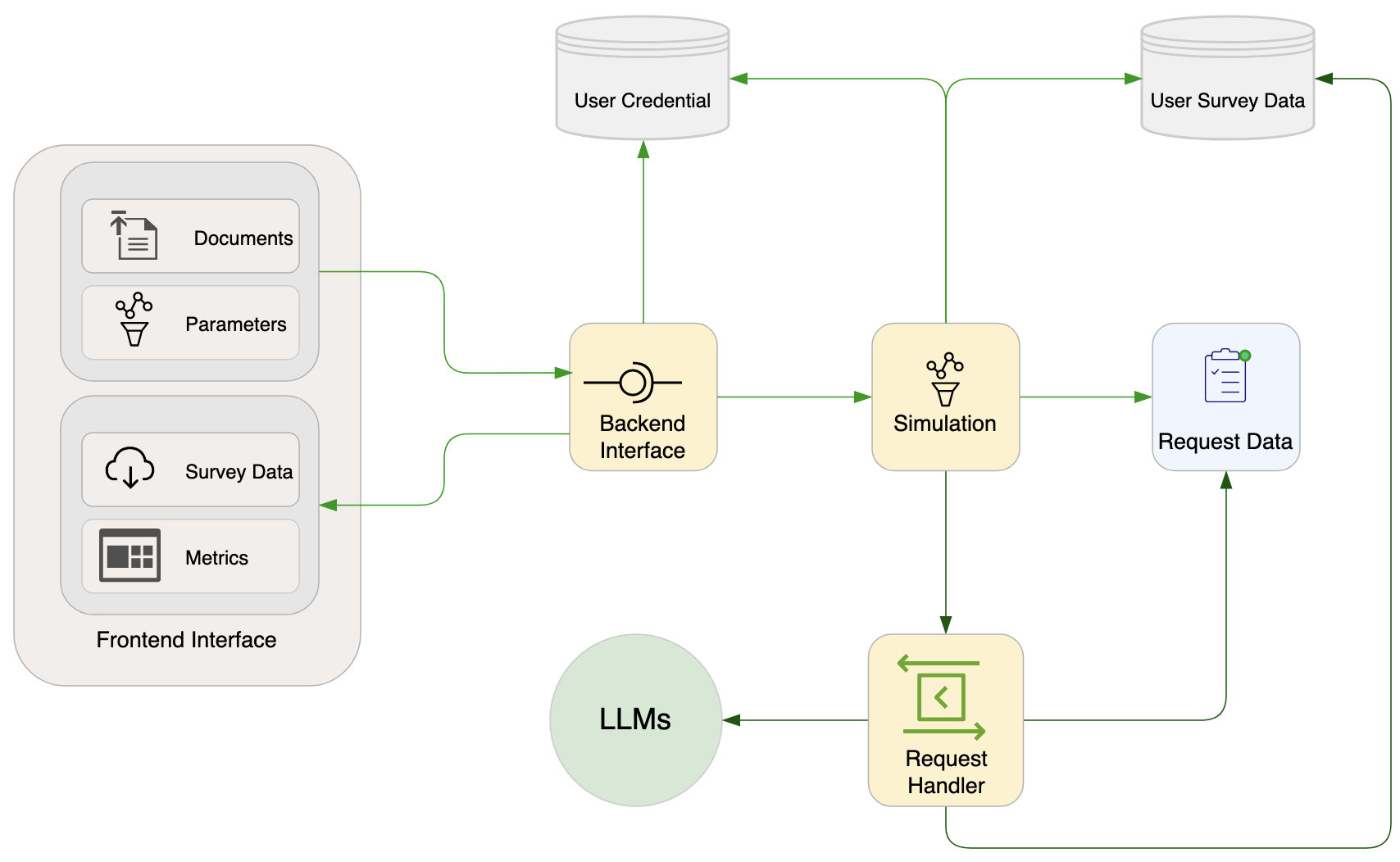}

}

\caption{Key components of the platform design}

\end{figure}

The platform's architecture consists of several key components designed
to enhance user interaction and facilitate research.

\textbf{Frontend Interface:} The interface serves as the gateway for
users to interact with the platform. It allows for easy parameter
configurations, document uploads, and provides real-time monitoring
metrics and a data viewer. The interface is designed to be user-friendly
and intuitive, minimizing the learning curve for new users.

\textbf{Databases:} Two types of databases are incorporated: one for
storing user credentials and another for per-user simulation data. This
separation is designed to enhance data security and privacy, as well as
to provide a smooth user experience by keeping all relevant data at
hand.

\textbf{Backend interface:} This is where a request is started and
ended. The backend interface is responsible for authenticating users,
processing simulation input data and sending survey data to the frontend
listener. It executes these tasks through appropriate interface calls to
the user credential database and the simulation component.

\textbf{Simulation:} This is where a survey process starts and ends. The
simulation component is responsible for processing uploaded documents,
constructing, creating request data, initialize the concurrent request
handler, and update metrics data for the frontend metrics listener. Each
agent possesses a profile constructed from parameter configurations and
a set of interfaces for interaction with the metrics monitor, databases,
and GPT models. The simulation component serves as a conduit, ensuring
efficient and error-free data flow within the platform. It also
internalizes the complexity of prompt engineering and related operations
needed to ensure that the LLMs behave consistently in term of the
structure of their response.

\textbf{Request data:} This component stores agent data and request data
in an optimal format created by the simulation component. It also serves
as a streaming data source to avoid unnecessary, and costly, memory
consumption that would take place while concurrent api calls are made
for each agent's request.

\textbf{Request Handler:} This component manages the complexity of
prompt engineering and concurrent API calls. It allows the simulation
process to communicate with the GPT models seamlessly and ensures all
prompts are correctly structured and successfully sent. It also manages
API calls to reduce the chance of hitting rate limits frequently while
ensuring that a simulation on large scale (e.g.~one with 1000 agents)
can run as fast as it could. Another important feature of this component
is the handling of retry failures due to rate limit and API request
errors. In the face of these failures, the handler will automatically
process the last safe survey results and return them in proper formats
for the agents to process further.

\hypertarget{sec-applications}{%
\section{Applications}\label{sec-applications}}

With the emergent capabilities of Large Language Models (LLMs) and the
ongoing development of Augmented Language Models (ALMs), understanding
the underlying premises that govern their behaviors and decision-making
processes becomes increasingly crucial, especially when they operate in
complex social settings and have real-world impact. This understanding
is essential to ensure the alignment of social AI with human values.

ALMs, which are essentially built on top of pre-trained LLMs like
OpenAI's gpt-4, incorporate various elements such as retrieval plug-ins,
different learning techniques (few-shot), diverse prompting methods
(such as chain-of-thought, self-model, and contextual prompts),
functional coding, and integration with other modalities like voice,
vision, and sound (Mialon et al., 2023). Additionally, future iterations
of ALMs are expected to incorporate different AI techniques such as
reinforcement learning and symbolic logic, enhancing their knowledge
organisation, reasoning, and learning capabilities.

Researchers have recognized the potential of LLMs as valuable tools to
study and probe the human mind and society (Arora et al., 2023; Binz \&
Schulz, 2023; Korinek, 2023; Miotto et al., 2022), given their training
on vast amounts of human data and their ability to generate human-like
text. Scholars have also discussed their potential in simulating human
subjects (Aher et al., 2023; Horton, 2023; Park et al., 2022).
Consequently, researchers from various disciplines such as behavioral
economics, cognitive psychology, social psychology, linguistics have now
started to investigate LLMs' behavior and decision-making processes and
its use as a scientific tool. However, the procedures and tuning of LLMs
(e.g., temperature, context window, prompt context and structure) for
judgment and evaluation of alignment are not yet standardized or
consistently applied across studies. Moreover, digital literacy and
programming skills continue to present significant obstacles for many
researchers to implement such research robustly at scale, particularly
those in behavioral economics and the social sciences.

Considering the fast-paced nature of research and development in AI at
the moment, it is essential to also extend our focus beyond pre-trained
LLMs and consider the emergent capabilities and value systems of ALMs
within various different social contexts. ALMs are increasingly
augmented with additional tools and various prompting techniques,
spanning different context windows and incorporating other modalities.
Furthermore, these ALMs are now actively performing real-world actions.
Calling a tool in the context of ALMs often involves having an impact on
the virtual or physical world and observing the resulting effects, which
are typically integrated into the ALM's ongoing context. Moreover, ALMs
are increasingly engaging in delegate actions such as carrying out
transactions on our behalf or responding to customer queries and emails
in human-like ways.

By acknowledging the advancements in ALMs and the complex nature of
their interactions with the world, we can gain a comprehensive
understanding of the premises underlying their behaviors and
decision-making processes by benchmarking through survey and
experimental methods. This knowledge is crucial for ensuring the
development of responsible and aligned social AI systems that reflect
human values for the benefit of all humankind.

SurveyLM facilitates this exploration in an easy and intuitive manner.
It empowers researchers to investigate the behaviors and decision-making
of LLMs and ALMs in a robust and systematic way, using an easy-to-use,
click-and-play online interface.

The SurveyLM platform is highly versatile and adaptable to a multitude
of decision-making scenarios and experimental settings. Here are just
some potential applications:

\begin{enumerate}
\def\labelenumi{\arabic{enumi}.}
\item
  \textbf{Survey Data Generation}: Simulate agents to answer an array of
  survey questions, creating a rich dataset that can mimic diverse,
  human-like responses to these questions. This can be particularly
  useful in preliminary research phases, hypothesis testing, or for
  enhancing existing datasets.
\item
  \textbf{Allocation Games}: Simulate scenarios where agents need to
  make decisions about resource allocation. This could involve public
  goods games, bargaining games, prisoner's dilemma scenarios, or other
  economic games that explore cooperation, competition, and negotiation.
\item
  \textbf{Cognitive Psychology Experiments and (Multi-Stage) Scenarios}:
  Simulate cognitive tasks and adaptation processes to understand
  decision-making processes, memory, attention, perception, and
  problem-solving strategies.
\item
  \textbf{Social Interaction Simulations}: Model and simulate complex
  social interactions within groups, examining phenomena like group
  dynamics, communication patterns, social influence, and conformity.
\item
  \textbf{Consumer Behavior Analysis}: Simulate buying decisions of
  agents to understand patterns in consumer behaviour, product
  preferences, and purchase rationales.
\item
  \textbf{Policy Impact Analysis}: Simulate reactions to new policies or
  regulations to gauge potential public response and impact (i.e., to
  understand emergent macro behaviors).
\item
  \textbf{Risk-taking (Multi-Stage) Scenarios}: Study decision-making
  under uncertainty or risk, such as in gambling or investment
  scenarios.
\item
  \textbf{Health-related Decision-making}: Explore choices related to
  health behaviors, preventative measures, treatment options, etc.
\item
  \textbf{Educational Settings}: Understand learning behavior, knowledge
  acquisition, and responses to different teaching methods.
\item
  \textbf{Environmental Decision-making}: Simulate agent decisions about
  resource usage, conservation behaviors, and responses to environmental
  policies.
\end{enumerate}

By simulating decision-making across a large spectrum of randomized
agent demographic attributes (e.g., age, gender, education level,
personality, etc.), SurveyLM provides a unique platform to investigate
even the most sensitive, challenging, or otherwise taboo subjects/topics
that are typically difficult to broach with human research participants,
such as sexuality, drug use or life-event shocks. By probing these areas
in a simulated environment, we leverage the potential of ALMs to explore
sensitive topics (e.g., health, social, economic, ethical, etc) in a
safe and ethical environment.

SurveyLM's potential is vast when it comes to exploring, e.g.,
sensitive, heated, challenging or taboo topics. Here are a few examples:

\begin{enumerate}
\def\labelenumi{\arabic{enumi}.}
\item
  \textbf{Mental Health}: Explore attitudes and behaviors around mental
  health issues, which are often stigmatized or misunderstood.
\item
  \textbf{Addiction}: Understand the complex dynamics of substance use
  and addiction, and the social attitudes towards these subjects.
\item
  \textbf{Sexuality}: Explore attitudes towards various sexual
  orientations, gender identities, or sexual behaviors.
\item
  \textbf{Religion and Faith}: Assess how individuals interact with
  religious beliefs and practices, including perceptions of other
  religions.
\item
  \textbf{Political Extremism}: Investigate the drivers of extreme
  political views, intolerance, or radicalisation.
\item
  \textbf{Race and Ethnicity}: Examine attitudes towards different races
  and ethnicities, including instances of bias, discrimination, and
  prejudice.
\item
  \textbf{Immigration}: Assess perceptions and misconceptions about
  immigration and immigrants.
\item
  \textbf{Body Image}: Explore attitudes and pressures around body image
  and physical appearance.
\item
  \textbf{(Economic) Inequality}: Understand perspectives on wealth
  distribution, poverty, and economic disparity or inequalities in
  general.
\item
  \textbf{Climate Change and Natural Disasters}: Investigate attitudes
  and beliefs about climate change and natural disasters, environmental
  responsibility, and sustainability.
\end{enumerate}

These topics are typically difficult to discuss openly, but SurveyLM
provides a secure and confidential platform for exploring them in depth.

We are confident that as SurveyLM evolves and adapts over time, it will
uncover numerous other promising areas of application. The examples
provided here represent just a fraction of the platform's potential,
highlighting only the possibilities we, and others in the field, have
identified to date. With the rapidly advancing landscape of social
science research, there is no doubt that the scope of SurveyLM's
application will continue to expand, revealing even more groundbreaking
possibilities in the future.

\hypertarget{sec-future-developments}{%
\section{Future Developments}\label{sec-future-developments}}

Despite its advanced design and capabilities, our research platform
faces certain limitations that we are actively seeking to address. These
hurdles present opportunities for enhancement and refinement,
contributing to the platform's ongoing evolution.

\textbf{Simplistic Agent Profile Configuration:} A significant
limitation lies in the current mechanistic profile configuration for
each agent (e.g., you are \textless AGE\textgreater, your personality is
\textless BIG 5 PROFILE\textgreater, and reside in
\textless LOCATION\textgreater). We do not draw on or attempt to
simulate human subjects from demographic backgrounds of past survey
respondents, as in e.g., (Argyle et al., 2023). Standard profile
constructs often used in survey studies underpin this design, providing
a simplified interaction model for users. However, these constructs'
rudimentary nature restricts the context within which the GPT models
function, sometimes compromising the depth and richness of their
responses. To capture the nuanced, multifaceted nature of human
contexts, we need a more sophisticated approach. The solution we are
developing and testing is a custom profile prompt feature, which will
enable users to create intricate, context-sensitive profiles for their
agents (e.g., profile construction by story telling). By broadening the
contextual basis of agent profiles, we anticipate an enhancement in the
model responses' relevance and applicability.

\textbf{OpenAI API Rate Limit Constraints:} The rate limits imposed by
OpenAI's API can influence the stability of output and latency, creating
potential bottlenecks for users requiring high-volume, real-time access
to the models while being able to receive responses to their complex
request in expected formats. A good solution depends on three things.
First, users must be able to obtain API keys that have the desirable
rate limits. Second, OpenAI, and other LLM providers, will increase or
phase out rate limits. Third, the platform's batching and request
mechanisms must be robust. We are currently enhancing our concurrent
request handler to optimize request scheduling and execution, thus
maximizing throughput within rate limits for a given API keys.
Additionally, we are in discussions with OpenAI to explore ways to
improve throughput and latency.

\textbf{Model Diversity:} Our platform currently supports only OpenAI's
models, chosen for their leading-edge capabilities and robust API
access. This model-specific dependency could limit the platform's
flexibility, as different models may offer unique strengths and
capabilities that could be beneficial in diverse research scenarios. We
are therefore actively testing other open source and commercial AI
models to potentially integrate into our platform (Bai et al., 2022;
Touvron et al., 2023), thus expanding its versatility and research
applicability. It should be noted that we avoid models that essentially
add new prompt-engineering layers on top of base ALMs to improve
decision performance in specific tasks (Shinn et al., 2023; Yao et al.,
2023).

\textbf{Realistic Condition Profiles:} A minor drawback of our platform
is the occasional generation of unrealistic agent profiles due to the
randomness of profile construction. This approach also means that
sometimes we end up with ``interesting'' agent profile combinations that
may seldom present in the real world (e.g., a male lesbian). While rare,
these cases can disrupt the research process and lead to unrealistic
model responses. One solution lies in conditional profile construction,
where agent attributes are selected based on real-world prevalence and
correlations. However, it is important to note that this randomness can
sometimes yield unique case studies that might not have been otherwise
considered, offering unexpected insights and interesting research
avenues.

\textbf{Multi-Agent Games and Interaction:} Currently, SurveyLM allows
for the simulation of an agent's participation in games with other
players only when the other players and interaction rules are hard-coded
into the uploaded questions and answer instruction. However, we are yet
to develop the capacity for interactions between different agents within
the same simulated population. To achieve this, a series of
sophisticated enhancements would be necessary. Key among these
improvements is the ability to form and manage groups of agents. These
groups function as independent social entities, engaging in intricate
social interactions within their own set boundaries. In this envisioned
application, users could define group sizes, roles within these groups
(e.g., the proposer and responder roles in the ultimatum game), and any
variations thereof. Furthermore, we would offer the ability to set
randomisation parameters for both roles and variations, adding yet
another layer of complexity and realism to the simulations. Another
captivating idea under this future avenue is the introduction of a chat
function between different agents. This feature would allow the
observation of direct communication patterns and language use within and
across agent groups, offering researchers another dimension to their
social agent studies.

In summary, while our platform faces certain limitations, we view these
as opportunities for growth and enhancement. Our commitment to
continuous development and user satisfaction drives us to persistently
explore innovative solutions to these challenges. This means that we are
open for feedback and suggestions. As we progress on this journey, we
look forward to unlocking further potential in facilitating complex
research through advanced ALM models.

For beta access and further details about the SurveyLM platform, please
contact the corresponding author of this paper.

\newpage{}

\hypertarget{references}{%
\section*{\texorpdfstring{\textbf{References}}{References}}\label{references}}
\addcontentsline{toc}{section}{\textbf{References}}

\hypertarget{refs}{}
\begin{CSLReferences}{1}{0}
\leavevmode\vadjust pre{\hypertarget{ref-aher_using_2023}{}}%
Aher, G., Arriaga, R. I., \& Kalai, A. T. (2023). \emph{Using large
language models to simulate multiple humans and replicate human subject
studies} ({arXiv}:2208.10264). {arXiv}.
\url{http://arxiv.org/abs/2208.10264}

\leavevmode\vadjust pre{\hypertarget{ref-argyle_ai_2023}{}}%
Argyle, L. P., Busby, E., Gubler, J., Bail, C., Howe, T., Rytting, C.,
\& Wingate, D. (2023). \emph{{AI} chat assistants can improve
conversations about divisive topics} ({arXiv}:2302.07268). {arXiv}.
\url{http://arxiv.org/abs/2302.07268}

\leavevmode\vadjust pre{\hypertarget{ref-arora_probing_2023}{}}%
Arora, A., Kaffee, L.-A., \& Augenstein, I. (2023). \emph{Probing
pre-trained language models for cross-cultural differences in values}
({arXiv}:2203.13722). {arXiv}. \url{http://arxiv.org/abs/2203.13722}

\leavevmode\vadjust pre{\hypertarget{ref-bai_constitutional_2022}{}}%
Bai, Y., Kadavath, S., Kundu, S., Askell, A., Kernion, J., Jones, A.,
Chen, A., Goldie, A., Mirhoseini, A., McKinnon, C., Chen, C., Olsson,
C., Olah, C., Hernandez, D., Drain, D., Ganguli, D., Li, D.,
Tran-Johnson, E., Perez, E., \ldots{} Kaplan, J. (2022).
\emph{Constitutional {AI}: Harmlessness from {AI} feedback}
({arXiv}:2212.08073). {arXiv}.
\url{https://doi.org/10.48550/arXiv.2212.08073}

\leavevmode\vadjust pre{\hypertarget{ref-binz_using_2023}{}}%
Binz, M., \& Schulz, E. (2023). Using cognitive psychology to understand
{GPT}-3. \emph{Proceedings of the National Academy of Sciences},
\emph{120}(6), e2218523120.
\url{https://doi.org/10.1073/pnas.2218523120}

\leavevmode\vadjust pre{\hypertarget{ref-haerpfer_world_2021}{}}%
Haerpfer, C., Inglehart, R., Moreno, A., Welzel, C., Kizilova, K.,
Diez-Medrano, J., \& Puranen. (2021). \emph{World values survey: Round
seven--country-pooled datafile.} {JD} Systems Institute \& {WVSA}
Secretariat, 7, 2021.

\leavevmode\vadjust pre{\hypertarget{ref-hofstede_cultures_2005}{}}%
Hofstede, G. (2005). Culture's recent consequences. \emph{Proceedings of
the Seventh International Workshop on Internationalisation of Products
and Systems}, 3--4.

\leavevmode\vadjust pre{\hypertarget{ref-horton_large_2023}{}}%
Horton, J. J. (2023). \emph{Large language models as simulated economic
agents: What can we learn from homo silicus?} ({arXiv}:2301.07543).
{arXiv}. \url{http://arxiv.org/abs/2301.07543}

\leavevmode\vadjust pre{\hypertarget{ref-korinek_language_2023-1}{}}%
Korinek, A. (2023). \emph{Language models and cognitive automation for
economic research} (30957). National Bureau of Economic Research.
\url{https://doi.org/10.3386/w30957}

\leavevmode\vadjust pre{\hypertarget{ref-mialon_augmented_2023}{}}%
Mialon, G., Dessì, R., Lomeli, M., Nalmpantis, C., Pasunuru, R.,
Raileanu, R., Rozière, B., Schick, T., Dwivedi-Yu, J., Celikyilmaz, A.,
Grave, E., LeCun, Y., \& Scialom, T. (2023). \emph{Augmented language
models: A survey} ({arXiv}:2302.07842). {arXiv}.
\url{https://doi.org/10.48550/arXiv.2302.07842}

\leavevmode\vadjust pre{\hypertarget{ref-miotto_who_2022}{}}%
Miotto, M., Rossberg, N., \& Kleinberg, B. (2022). \emph{Who is {GPT}-3?
An exploration of personality, values and demographics}
({arXiv}:2209.14338). {arXiv}. \url{http://arxiv.org/abs/2209.14338}

\leavevmode\vadjust pre{\hypertarget{ref-park_social_2022}{}}%
Park, J. S., Popowski, L., Cai, C. J., Morris, M. R., Liang, P., \&
Bernstein, M. S. (2022). \emph{Social simulacra: Creating populated
prototypes for social computing systems} ({arXiv}:2208.04024). {arXiv}.
\url{http://arxiv.org/abs/2208.04024}

\leavevmode\vadjust pre{\hypertarget{ref-shinn_reflexion_2023}{}}%
Shinn, N., Cassano, F., Labash, B., Gopinath, A., Narasimhan, K., \&
Yao, S. (2023). \emph{Reflexion: Language agents with verbal
reinforcement learning} ({arXiv}:2303.11366). {arXiv}.
\url{https://doi.org/10.48550/arXiv.2303.11366}

\leavevmode\vadjust pre{\hypertarget{ref-shridhar_alfworld_2021}{}}%
Shridhar, M., Yuan, X., Côté, M.-A., Bisk, Y., Trischler, A., \&
Hausknecht, M. (2021). \emph{{ALFWorld}: Aligning text and embodied
environments for interactive learning} ({arXiv}:2010.03768). {arXiv}.
\url{https://doi.org/10.48550/arXiv.2010.03768}

\leavevmode\vadjust pre{\hypertarget{ref-tamkin_understanding_2021}{}}%
Tamkin, A., Brundage, M., Clark, J., \& Ganguli, D. (2021).
\emph{Understanding the capabilities, limitations, and societal impact
of large language models} ({arXiv}:2102.02503). {arXiv}.
\url{https://doi.org/10.48550/arXiv.2102.02503}

\leavevmode\vadjust pre{\hypertarget{ref-touvron_llama_2023}{}}%
Touvron, H., Martin, L., Stone, K., Albert, P., Almahairi, A., Babaei,
Y., Bashlykov, N., Batra, S., Bhargava, P., Bhosale, S., Bikel, D.,
Blecher, L., Ferrer, C. C., Chen, M., Cucurull, G., Esiobu, D.,
Fernandes, J., Fu, J., Fu, W., \ldots{} Scialom, T. (2023). \emph{Llama
2: Open foundation and fine-tuned chat models} ({arXiv}:2307.09288).
{arXiv}. \url{https://doi.org/10.48550/arXiv.2307.09288}

\leavevmode\vadjust pre{\hypertarget{ref-yao_react_2023}{}}%
Yao, S., Zhao, J., Yu, D., Du, N., Shafran, I., Narasimhan, K., \& Cao,
Y. (2023). \emph{{ReAct}: Synergizing reasoning and acting in language
models} ({arXiv}:2210.03629). {arXiv}.
\url{https://doi.org/10.48550/arXiv.2210.03629}

\end{CSLReferences}

\end{document}